\documentclass{egpubl}
\usepackage{pg2019}
 
\SpecialIssueSubmission    

\usepackage[T1]{fontenc}
\usepackage{dfadobe}  

\BibtexOrBiblatex
\electronicVersion
\PrintedOrElectronic

\usepackage[utf8]{inputenc} 
\usepackage[T1]{fontenc}    
\usepackage{url}            
\usepackage{booktabs}       
\usepackage{amsfonts}       
\usepackage{nicefrac}       
\usepackage{microtype}      

\usepackage{times}
\usepackage{epsfig}
\usepackage{graphicx}
\usepackage{amsmath}
\usepackage{amssymb}
\usepackage{comment}

\DeclareOldFontCommand{\bf}{\normalfont\bfseries}{\mathbf}
\DeclareOldFontCommand{\it}{\normalfont\itshape}{\mathit}
\DeclareOldFontCommand{\rm}{\normalfont\rmfamily}{\mathrm}
\DeclareOldFontCommand{\cal}{\normalfont\calseries}{\mathcal}

\title[Deform-and-learn]%
{Learning Body Shape and Pose from Dense Correspondences}

\author[Y. Yoshiyasu \& L. Gamez]
{\parbox{\textwidth}{\centering Y. Yoshiyasu and L. Gamez}\\ CNRS-AIST JRL}
	
\begin{document}
	
	
	\maketitle
	
	

	\begin{abstract}
		In this paper, we address the problem of learning 3D human pose and body shape from 2D image dataset, without having to use 3D dataset (body shape and pose). The idea is to use dense correspondences between image points and a body surface, which can be annotated on in-the wild 2D images, and extract and aggregate 3D information from them. To do so, we propose a training strategy called ``deform-and-learn" where we alternate deformable surface registration and training of deep convolutional neural networks (ConvNets). 		Unlike previous approaches, our method does not require 3D pose annotations from a motion capture (MoCap)  system or human intervention to validate 3D pose annotations.
	\end{abstract}

	


\section{Introduction}

Estimating 3D human pose and body shape from a single image is a challenging yet important problem, with a wide variety of applications such as computer animation and virtual try-on in fashion. 

Capturing and modeling of 3D body shape and pose has been mostly done in controlled settings using specialized 3D scanners such as a whole-body laser range scanner and motion capture (MoCap) system. With the progress of deep convolutional neural networks (ConvNets), 3D body shape can be estimated from a single image by regressing the parameters of statistical human body models. Most of current methods rely on 3D database for both body shape and pose, which still requires expensive 3D scanning systems to construct and extend their dataset that they use for training. 

Nonetheless, the capability of the current methods for expressing body shape and pose is rather limited because of two main reasons. Firstly, regression of body shape parameters is inherently difficult for deep ConvNets. The mapping between the input image and parameters of statistical body models is highly nonlinear and is currently difficult to learn. The second challenge is the lack of a large-scale 3D dataset. In fact, most of 3D body shape dataset are limited to the age range from young adult to middle age. Also, MoCap dataset for 3D human pose estimation are limited to a small variety of subjects, since it needs a complicated experimental setup where MoCap and RGB video cameras have to be synchronized. Because those motion data are acquired in a controlled environment, they are somewhat different from the natural poses that can be found in the in-the-wild images.   



{\it ``Can we learn 3D human body shape and pose directly from 2D images?''} In this paper, we tackle this challenging problem to bypass the 3D dataset scarcity problem by extracting and aggregating 3D information from dense correspondences annotated on 2D images. We propose a strategy called ``deform-and-learn" where we alternates deformable surface registration that fits a 3D model to  2D images and training of deep neural network that predicts 3D body shape and pose from a single image. Given dense image-to-surface  correspondences, the first registration step fits a template model to images. The result is then used as supervisional signals of 3D body shape and pose for training deep ConvNets in the second registration step. These two processes are iterated to improve accuracy. Unlike previous approaches, our method does not require statistical body shape models, 3D pose annotations from MoCap dataset or human interventions to validate 3D pose. 

The contributions of this paper are summarized as follows:

\begin{itemize}
	\item 
	We propose a deform-and-learn training strategy that alternates deformable registration and training of a deep ConvNets for estimating human body shape and pose. It uses dense correspondences annotated on 2D images, without having to use 3D dataset (body shape and pose).
	
	\item 
	To design a pose prior from 2D pose dataset, we propose a conditional generative adversarial networks (cGANs) for detecting 3D human joint positions from 2D keypoints. We incorporate geometric constraints in cGANs to further constrain 3D human pose predictions. The results are used as soft constraints to guide the training of deep ConvNets for body shape and pose estimation. 
	
	\item 
	We propose a skeleton-based deformable registration technique using back propagation, which can be implemented using a deep learning framework and parallelized with GPU. With the autograd technique, adding and minimization of various types of losses can be made simple, which frees our method from relying on 3D dataset and pre-built 3D statistical models. 
	
	\item 
	We propose a deep ConvNets that predicts body shape and pose using scalings of body segments as   body shape representation. With the final refinement step based on deformable registration using dense correspondence predictions, we can further align a body mesh model to an image.

\end{itemize}

\begin{figure*}[h]
	\centering
	\includegraphics[width=1\linewidth]{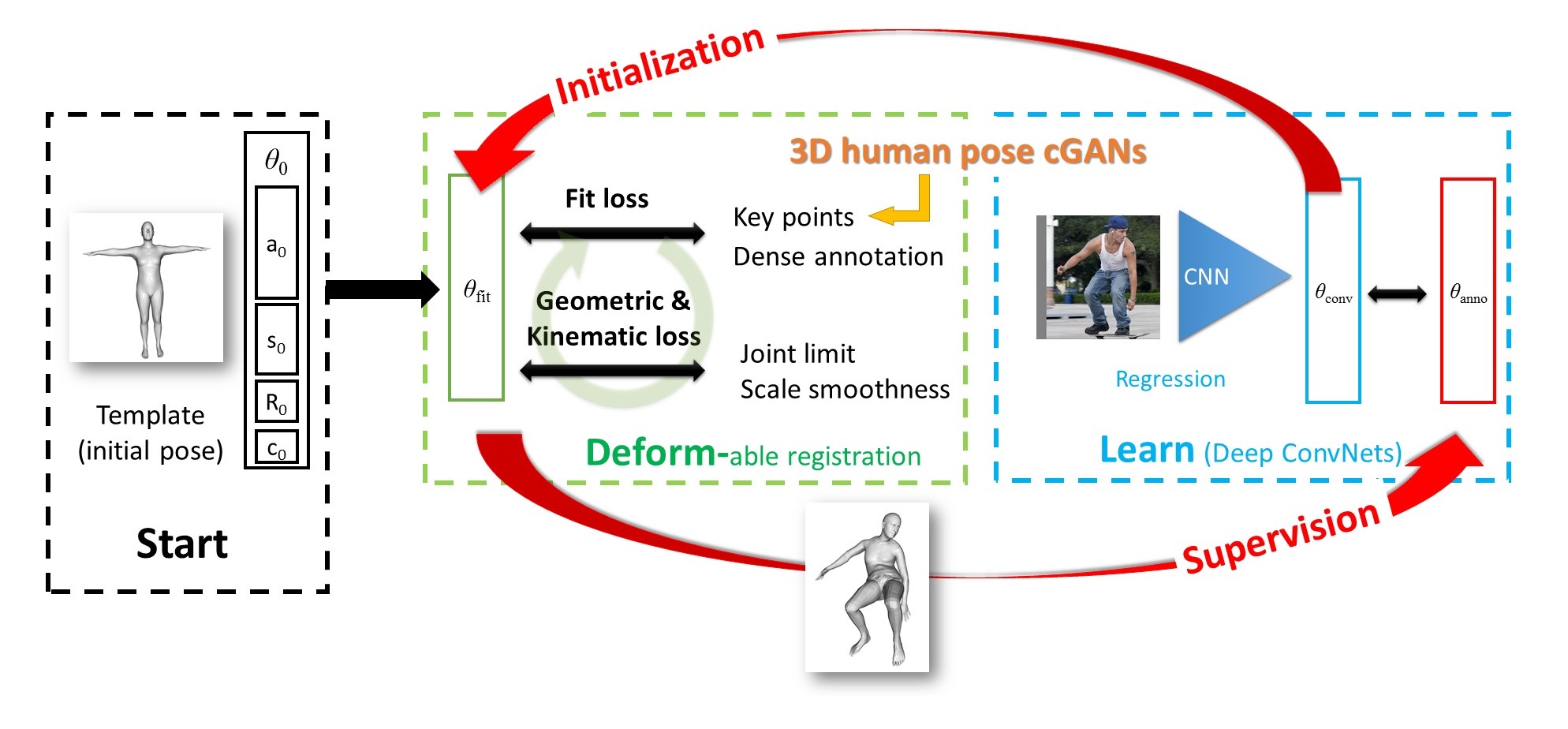}
	\caption{Overview of our deform-and-learn training strategy that iteratively performs deformable registration and deep learning. Let $\mathbf{\theta}$ be the parameters of the body model, such as body shape and pose. In the very beginning, the initial pose of registration is in the T-pose, ${\bf \theta}_0$. Given dense image-to-surface  correspondences, the first registration step fits a template model to images. After registration, we obtain a collection of ${\bf \theta}_{\rm fit}$ which is then used as supervisional signals ${\bf \theta}_{\rm anno}$ to train deep ConvNets that predicts body parameters ${\bf \theta}_{\rm conv}$. The results of ConvNets are used as  initial poses of deformable registration in the next round. These two steps are iterated to get better results.}
	\label{fig:overview}
\end{figure*}

	\section{Related Work}

\noindent {\bf Human body shape modeling and surface registration } Previously, modeling of 3D body shape is done with 3D scanners. The first approach in this line of work is done by Allen et al. \cite{ACP03} where the authors fit a template 3D body model to Caesar dataset that contains a couple of thousand subjects and used principal component analysis (PCA) to model the space of human body shape. Later, several techniques are proposed to extend the method of Allen et al. to handle both body shape and pose variations (such as SCAPE  \cite{ASK05} and SMPL \cite{bogo2016keep}) and even dynamic deformations (eg. Dyna \cite{Dyna:SIGGRAPH:2015}). Nonrigid  surface registration techniques have been used in body shape modeling to fit a template mesh models to 3D scans \cite{HAW08,ACP03, SP04, ARV07, YLSL10}.


\noindent {\bf Estimating 3D joint positions from an image } Early approaches predict 3D joint positions from key points by assuming that the almost perfect 2D key points are already extracted from an image \cite{ramakrishna2012reconstructing}. The first method based on ConvNets directly regresses 3D joint positions with an image \cite{li20143d}. Recent techniques achieves higher accuracy with an end-to-end framework that predicts 2D joints with heatmaps and then regresses 3D joint positions or depths from them \cite{zhou2017weakly, mehta2016monocular}. Martinez et al. \cite{martinez_2017_3dbaseline} on the other hand proposed a very simple network architecture that maps 2D joint coordinates to 3D joint positions, resulting in a two separate networks which can also achieve high accuracy. Pavlakos et al. \cite{PavlakosZDD16} used a volumetric heatmap representation which is a ConvNet friendly representation and can avoid regressing the real values in a highly nonlinear manner. Some methods regress kinematic parameters \cite{zhou2016deep, Skeleton_Yoshiyasu2018} to preserve human skeletal structures.

\noindent {\bf Body shape from an image } A common way to predict 3D human body shape and pose from an image is to employ pre-built statistical human models. Guan et al. \cite{Guan:ICCV:2009} first manually extract 2D keypoints and silhouettes of human body. The first automatic method was proposed in SMPLify \cite{bogo2016keep} where the human statistical model called SMPL was fitted to the 2D keypoints estimated from an image using ConvNets by an optimization technique. Tan et al. \cite{TanBC17} proposed an  indirect approach to learn body shape and pose by minimizing the estimated and real silhouettes. Tung et al. \cite{tung2017self} proposed a self-supervised learning motion capture technique that optimizes SMPL body parameters and Kanazawa et al. \cite{hmrKanazawa17} proposed an end-to-end learning system of human body and shape based on generative adversarial networks (GANs). More recently, silhouettes \cite{pavlakos2018humanshape,varol18_bodynet} and part segmentations \cite{NBF:3DV:2018} are incorporated to improve prediction accuracy. On the other hand, in DensePose \cite{Guler2018DensePose} uv coordinates and part indices are directly annotated on images to establish image-to surface dense correspondences but this is still not complete 3D representation. The most similar approach to ours would be Lassner et al. \cite{Lassner:UP:2017} where the authors proposed a method to construct a 3D human  body shape and pose dataset by fitting a SMPL model to images. Compared to them, our  approach does not require statistical 3D pose/shape priors or human interventions to validate pose fits.

\noindent {\bf Unsupervised and weakly-supervised approaches } Given 2D keypoints of human joints, Kudo et al. \cite{kudo2018} and Chen et al. \cite{Chen2019} used conditional generative adversarial networks (GANs) to estimate 3D human pose only from 2D pose dataset. Rodin et al. \cite{rhodin2018unsupervised} used auto-encoder to compress multi-view images into latent variables and they reconstructed 3D human pose from them, which does not need a large amount of 3D pose supervisions. 
   




	\section{Problem formulation}

The goal of our work is to learn a model that can predict 3D body shape and pose from a single image using deep ConvNets, without having to use 3D dataset. To the best of our knowledge, this paper is the first one to achieve it. To that end, we use dense correspondence annotations (Fig. \ref{fig:dennse_annotation}) between image points and a body surface, which can be annotated on 2D images in-the-wild and provides rich information about body shape and pose. Compared to silhouettes and part segmentation, dense correspondence annotations are less noisy around boundaries and can be obtained with some more additional human efforts whose annotation time is almost the same as that of part segmentation  \cite{Guler2018DensePose}. 

Although dense correspondences between a body surface and image points contain rich information, they themselves are not sufficient for recovering 3D body shape and pose, especially for depth. The strategy we take in this paper is to incorporate geometric and kinematic losses imposed on body parameters as well as an adversarial loss defined from 2D key points to constrain the space of body shape and pose, as we do not have a direct access to 3D dataset that we can use for training. Consequently, the total loss we define and wish to minimize is as follows: 
\begin{align}
{\cal L}  =  {a}{\cal L}_{\rm dense} + {b}{\cal L}_{\rm geo} + {c}{\cal L}_{\rm adv}
\label{eq:loss}
\end{align}
where ${\cal L}_{\rm dense}$ is the dense correspondence loss which penalizes the inconsistency of fits between the body model and images defined in terms of dense correspondences, ${\cal L}_{\rm geo}$ is the geometric and kinematic loss for regularization and ${\cal L}_{\rm adv}$ is the adversarial loss to constrain the distribution of the predicted poses close to that of 2D keypoint annotations. The weights  $a$, $b$ and $c$ control the relative strengths of the terms.

\noindent {\bf Deform-and-learn iterative training strategy } Directly minimizing all of the losses in Eq. (\ref{eq:loss}) at the same time is difficult and in fact we experienced that the error  stayed high. Instead, we decouple the problem into three components: A) training of a conditional generative adversarial networks that predicts 3D joints from 2D keypoints; B) optimization of latent body parameters based on image-surface registration; C) learning of body parameters by providing latent supervisions obtained in step B. The first component is trained once and provides soft constraints of 3D joints in the second and third step. The second and third components are iterated for several times to improve the accuracy, which we refer to as the ``deform-and-learn'' training strategy. 

We found that decoupling of the training phase into three steps and providing supervisions on latent variables works effective. In fact, recent approaches showed that providing supervisions on latent body parameters are effective in stabilizing and improving training \cite{NBF:3DV:2018}. Here we reconstruct latent variables from dense annotations by deformable surface registration. Note that deformable registration is an local optimizer and is sensitive to an initial solution. This is why we propose the iterative training strategy ``deform-and-learn'', where we alternate between deformable registration and learning. This strategy will gradually improve performance by updating the initial solution of the registration phase and then the latent supervisions in the learning phase.

\noindent {\bf Body shape and pose model } To fit a template mesh model to an image, we use a skeleton-based parametric deformable model which is a modified version of SMPL  \cite{bogo2016keep}. The template mesh consists of $n$ vertices, where the number of vertex $n$ is 6980 in this paper. The vertex positions of the template, $\mathbf{v}_1 \ldots \mathbf{v}_n$, are denoted by a $n \times 3$ vector, $\mathbf{v}= [\mathbf{v}_1 \ldots \mathbf{v}_n]^\mathrm{ T}$. The pose of the body is defined by a skeleton rig with 23 joints where the pose parameters ${\bf a} \in \mathbb{R}^{24 \times 3}$ is defined by the axis angle representation of the relative rotation between segments. The body model is posed by a joint parameters ${\bf a}$ via forward kinematics. Instead of using a low-dimensional shape space as in \cite{bogo2016keep}, which can be learned from thousands of registered 3D scans, we use segment scales to model a body shape, which is parametrized by segment scales ${\bf S} \in \mathbb{R}^{24}$. This way, body shape can be modeled more flexibly without the need to use 3D body scans---it does not have to be confined in the space of statistical models. Using linear blend skinning, the body deformation model is defined as a function ${\bf v} = X({\bf S}, {\bf a})$. 

We use the weak-perspective camera model and solve for the global rotation ${\bf R} \in \mathbb{R}^{3 \times 3}$, translation  ${\bf t} \in \mathbb{R}^{2}$ and global scale $s \in \mathbb{R}$. Rather than using other rotational representation such as axis angle, we directly optimize for a rotation matrix with 9 parameters due to its property to represent orientations uniquely in 3D space.  Since this approach makes a transformation deviating from a rotation matrix, we applied the Gram Schmidt normalization to ortho-normalize the matrix. Thus the total number of the parameters representing human body is 108,  ${\bf \theta} = [{\bf a}, {\bf S},{\bf R}, s, {\bf t}]$. With the body parameters ${\bf \theta}$, deformation and projection of vertices ${\bf v} = X({\bf S}, {\bf a})$ into an image is achieved as:
\begin{align}
{\bf x} = s \Pi ({\bf R} X({\bf S}, {\bf a})) + {\bf t}  \nonumber
\end{align}
where $\Pi$ is an  orthogonal projection.

\section{Overview}
The overview of our approach is depicted in Fig. \ref{fig:overview}. We train a conditional generative adversarial networks (cGANs) that predicts 3D joint positions from 2D joint positions, which will guide the registration and training of deep ConvNets for body shape and pose (Section \ref{sec:cgan}). The deform-and-learn training strategy alternates deformable surface registration that fits a 3D model to 2D images and training of deep neural network that predicts 3D body shape and pose from a single image (Section \ref{sec:regist}). In the very beginning, the initial pose of registration is in the T-pose, ${\bf \theta}_0$. Given image-surface dense correspondences, the first registration step fits a template model to images. After registration, we obtain a collection of body parameters ${\bf \theta}_{\rm fit}$ which is then used as supervisional signals ${\bf \theta}_{\rm anno}$ in order to train deep ConvNets that predicts body parameters ${\bf \theta}_{\rm conv}$ (Section \ref{sec:conv}). The results are used as  initial poses of surface registration in the next round. This training process is iterated for several times to get better results.

In the inference phase, we optionally perform refinement based on deformable surface registration. We first use the trained deep ConvNets to predict body shape and pose parameters, we refine the result using the registration technique starting from the ConvNet result as an initial solution. Thus the overall component used here is the same as the training phase, except that the order is flipped.

	\begin{figure*}[tb]
	\centering
	\includegraphics[width=0.9\linewidth]{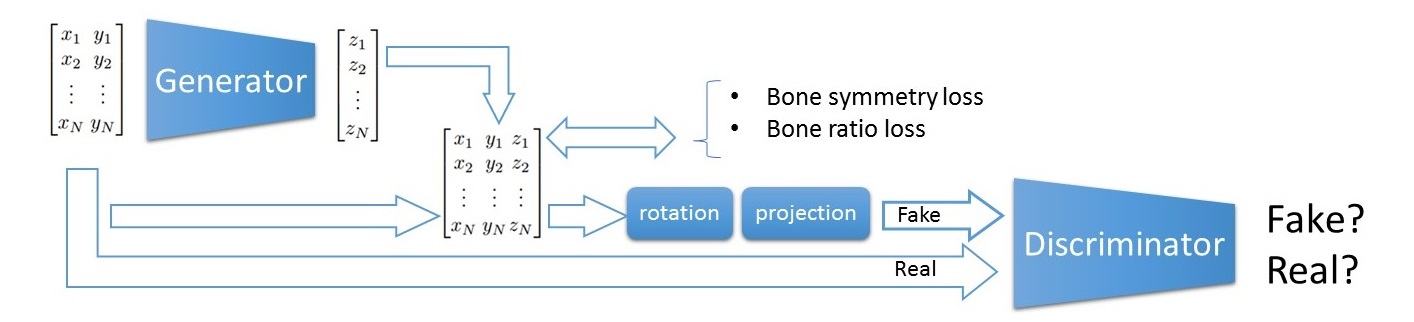}
	\caption{3D human pose cGANs with geometric constraints. The input to the generator is the 2D key points of $N$ joints and the output is depths of those joints. Once the generator outputs the depths values $z_i$, they are concatenated with $x_i$ and $y_i$ coordinates. This 3D joint position is rotated about the vertical axis and projected to the image space. The discriminator inputs the projected joint positions as $fake$ and the 2D keypoint data as $real$. In addition to the adversarial loss, we incorporate geometric constraints, such as bone symmetry constraints, to further constrain the space of solution.}
	\label{fig:GAN}
\end{figure*}

\section{Conditional generative adversarial networks for 3D human pose with geometric constraints}
\label{sec:cgan}

We propose a conditional generative adversarial networks (GANs) to predict depths of joints from 2D keypoints in an unsupervised manner. The results of the generator is used as soft constraints to guide image-surface registration in the next section. 

We take a similar approach as Kudo et al. \cite{kudo2018} and Chen et al. \cite{Chen2019} where the 3D joint positions produced by a generator network ($G$) is projected to the image plane to obtain 2D joint positions and a discriminator ($D$) judges real or fake in 2D image space. The key difference of our model from previous approaches \cite{kudo2018} is that our approach incorporates geometric constraints, such as bone symmetry constraints, to further constrain the space of solution. The network architecture is depicted in Fig. \ref{fig:GAN}. The input to the generator is the 2D key points of $N$ joints and the output is depths of those joints. The predicted depths values $z_i$ are then  concatenated with $x_i$ and $y_i$ coordinates, rotated around the vertical axis and projected to the image space. The discriminator inputs the projected joint positions as $fake$ and the 2D keypoint data as $real$. For both networks, we use multi-layer perceptron (MLP) with eight linear layers to map 2D coordinates to depths and binary class. 

Let ${\bf u}$ be the 2D joint positions of a skeleton. Also let us denote an angle around the vertical axis as $\phi$. Our 3D human pose cGANs uses the following standard adversarial loss functions for $G$ and $D$: 
\begin{align}
{\cal L}_{\rm adv}^G &= E_{{\bf u}, \phi} [\log (1- D (f({\bf u}, G({\bf u}); \phi)))  \nonumber \\ {\cal L}_{\rm adv}^D &= E[\log D({\bf u})] \nonumber 
\end{align}
where $f$ denotes the rotation and the projection function. Note that we validate the pose from multiple views where we used angles [deg], $\phi =  \{45, 60, 90, 135, 180, 235, 270 \}$ for each pose. 

In addition to the adversarial loss, the geometric loss is also applied. Specifically, we use the bone symmetry loss ${\cal L}_{\rm sym}$ that constrain the left and right limb be similar and the bone ratio loss ${\cal L}_{\rm ratio}$ that minimizes the difference between the normalized bone length prediction and that of dataset. The bone ratio loss ${\cal L}_{\rm ratio}$ is defined as: 
\begin{equation}
{\cal L}_{\rm ratio} = \sum_{e \in{\cal B}} \|\frac{l_e}{l_{\rm trunk}} - \frac{\bar{l}_e}{\bar{l}_{\rm trunk}}   \|^2 \nonumber
\end{equation}
where $\frac{l_e}{l_{\rm trunk}}$ is the ratio of the bone length for bone $e$ in a set of bones  ${\cal B}$ in a skeleton   with respect to the trunk length and $\frac{\bar{l}_e}{\bar{l}_{\rm trunk}}$ is that of the average skeleton.   Let ${\cal B}_s$ be the set of symmetry pairs of bone segments which contains indices of bones e.g., the left and right forearm. Then the bone symmetry loss ${\cal L}_{\rm sym}$ is defined as: 
\begin{equation}
{\cal L}_{\rm sym} = \sum_{i,j \in{\cal B}_s } \|l_i- l_j \|^2 \nonumber\\
\end{equation}
where $l_i$ and $l_j$ is the lengths of the bone for symmetry bone pairs. We mix the above losses to train the generator such that the loss is:
\begin{equation}
{\cal L}^G = \epsilon{\cal L}_{\rm adv}^G + ({\cal L}_{\rm ratio} + {\cal L}_{\rm sym}) \nonumber\\
\end{equation}
where $\epsilon$ is the weight for controlling the strength of the adversarial term, which we set to 0.1 in this paper.

\section{Image-surface deformable registration} 
\label{sec:regist}
We propose a deformable surface registration technique to fit a template mesh model to images to obtain 3D body shape and pose annotations for training deep ConvNets. Here deformable  registration is formulated as a gradient-based method based on back propagations, which can be implemented with a deep learning framework and parallelized with GPU. With the automatic differentiation mechanisms provided with a deep learning framework, adding and minimizing various kinds of losses have made easy and straightforward. As a result, the proposed deformable registration technique thus incorporates kinematic, geometric and correspondence losses. 

Given image-surface dense correspondences annotated on images, the template mesh is fitted to images by optimizing body parameters ${\bf \theta} = [{\bf a}, {\bf S},{\bf R}, s, {\bf t}]$ subject to kinematic and geometric constraints. In total, the overall loss function for our  registration is of the form:
\begin{align}
{\cal L}_{\rm regist}  &= \omega_{\rm dense} {\cal L}_{\rm dense} + \omega_{\rm KP} {\cal L}_{\rm KP} \\ \nonumber &+ \omega_{\rm scale} {\cal L}_{\rm scale} + \omega_{\rm joint} {\cal L}_{\rm joint} \nonumber + \omega_{\rm det} {\cal L}_{\rm det} \nonumber 
\end{align}
where ${\cal L}_{\rm dense}$ and ${\cal L}_{\rm KP}$ are the dense correspondence and key point losses that penalize the alignment inconsistency of the body model and images defined in terms of dense correspondences and key points. The losses ${\cal L}_{\rm  scale}$ and ${\cal L}_{\rm  joint}$ is the segment scaling smoothness and kinematic loss for regularization. The transformation determinant loss ${\cal L}_{\rm det}$ makes the determinant of the global transformation positive. In addition,  $\omega_{\rm dense}$,  $\omega_{\rm KP}$, $\omega_{\rm scale}$, $\omega_{\rm joint}$ and $\omega_{\rm det}$ are the respective weights for the above defined losses. The initialization of body parameters is provided from the predictions of deep ConvNets. For the very first iteration where the Convnet predictions are not available, segment scale ${\bf S}$ is set 1 for all segments and pose ${\bf a}$ is set to 0 for all joints, which means that registration is started from the T pose.




\subsection{Correspondence fit loss}
\label{sec:corresp}

The correspondence loss comprises two losses: the dense correspondence loss ${\cal L}_{\rm Dense}$ and keypoint loss ${\cal L}_{\rm KP}$. 

\noindent {\bf Dense correspondence loss } Let us define a set of image-surface correspondences $\mathcal{C}=\{(\mathbf{p}_{1}, \mathbf{v}_{\mathrm{idx}(1)}) \ldots (\mathbf{p}_{N}, \mathbf{v}_{\mathrm{idx}(N)}) \}$, where $\mathbf{p}$ is the image points. In addition $\mathrm{idx}(i)$ is the index of the mesh vertices that is matched with image point $i$. Now we can define the dense correspondence loss as:
\begin{eqnarray}
\label{eq:Ec}
{\cal L}_{\rm dense} = \sum_{i\in \mathcal{C}} \|{\bf p}_{i} - \mathbf{x}_{\mathrm{idx}(i)} \|^2 \nonumber
\end{eqnarray}
where a mean squared error (MSE) is used to calculate the loss.

\noindent {\bf Key point loss } To produce 3D poses with statistically valid depths, the results of cGAN is used to guide deformable registration. Instead of attaching a discriminator to the registration framework, the depth values from cGAN and the ground truth 2D joint coordinates are provided as a soft constraint to constrain the position of the 3D joints based on the MSE loss:
\begin{eqnarray}
{\cal L}_{\rm KP} = \sum_{i\in \mathcal{J}} \|x_i - \bar{x}_i \|^2 
+ \sum_{i\in \mathcal{J}} \|y_i - \bar{y}_i \|^2
+ \sum_{i\in \mathcal{J}} \|z_i - z_i^{\rm GAN} \|^2 \nonumber
\end{eqnarray}
where $\bar{x}_i$ and $\bar{y}_i$ are the ground truth of 2D key points. Also $z_i^{\rm GAN}$ is the depth at joint $i$ predicted by cGANs. 


\subsection{Geometric and kinematic loss}
Since we attract the template mesh to 2D image coordinates, the problem is ill-posed and deformations are not constrained. Thus we introduce the regularization terms that avoids extreme deformations. 

\noindent {\bf Segment scaling smoothness } To avoid extreme segment scalings, we introduce the scaling smoothness loss, which minimizes difference between scalings of adjacent segments:
\begin{equation}
{\cal L}_{\rm scale} = \sum_{e \in \mathcal{B}} \|{\bf S}_e - {\bf S}_{{\rm adj}(e)}\|^2 \nonumber
\end{equation}  
\noindent {\bf Joint angle smoothness and limit loss } To prevent extreme poses, we introduce joint angle smoothness loss and joint limit loss. The joint smoothness loss is enforced at every joint in a skeleton, $\mathcal{J}$, and will contribute to avoid extreme bending. To avoid hyper-extensions which will bend certain joints like the elbows and knees (where we represent as $\mathcal{J'}$) in the negative direction, we introduce the joint limit loss. The regularizations that act on joints are thus represented as:
\begin{equation}
{\cal L}_{\rm joint} =  \sum_{i \in \mathcal{J}} \|	{\bf a}_i\|^2 + \sum_{i \in \mathcal{J'}} \|{\rm exp}({\bf a}_i)\|^2 \nonumber
\end{equation}  
where the first term minimizes joint angles whereas the latter term penalizes rotations violating natural constraints by taking exponential and minimizing it. 

\noindent {\bf Transformation determinant loss } Since we use a rotation matrix for representing the global rotation at the root, it is necessary to apply a constraint on a matrix to keep its determinant to positive. Thus, we define the transformation determinant loss as: 
\begin{eqnarray}
{\cal L}_{\rm det} =  \exp (-{\rm det}({\bf R})) \nonumber
\end{eqnarray}


\begin{figure*}[t]
	\centering
	\includegraphics[width=1\linewidth]{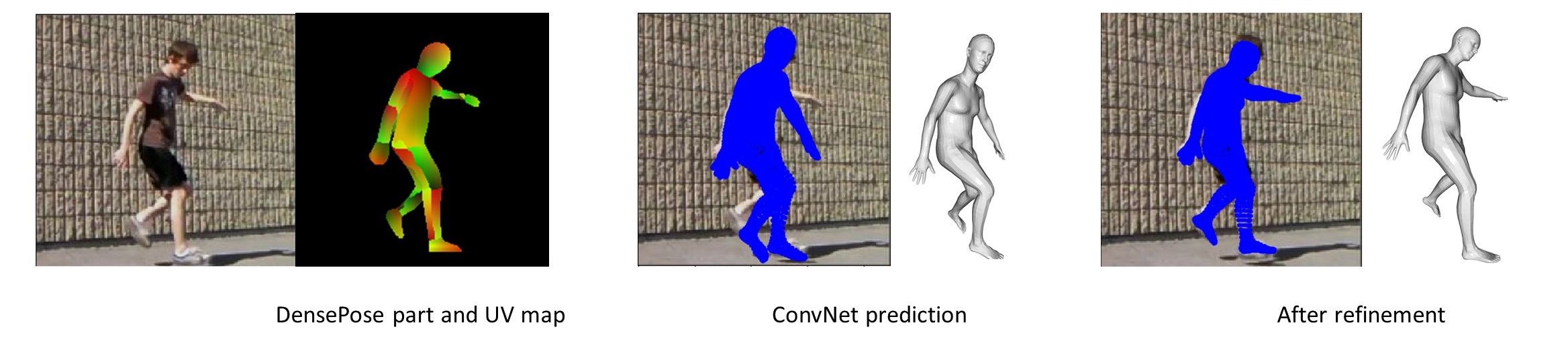}
	\caption{Refinement based on densepose prediction. The result of body parameter regression networks is refined based on a deformable surface registration technique. To define the dense correspondence term, we converted to image-surface correspondences obtained from DensePose predictions of uv maps and part indices. }
	\label{fig:refine_method}
\end{figure*}

\section{Estimating 3D human body shape and pose from a single image}
\label{sec:conv}
\subsection{Deep ConvNets for body shape and pose regression}

Using the results obtained by deformable registration as annotations for training deep ConvNets, we regress body shape and pose parameters with an image. We also add the dense correspondence and keypoint losses as in Section \ref{sec:corresp} for additional supervisions. In total, we minimize the loss function of the form:
\begin{equation}
{\cal L}_{\rm conv} =  \alpha {\cal L}_{\rm regress} +   \beta {\cal L}_{\rm dense} + \gamma {\cal L}_{\rm KP}  \nonumber
\end{equation}
where ${\cal L}_{\rm regress}$ is the regression loss for body parameters. $\alpha$, $\beta$ and $ \gamma$  are the respective weights. Let $\theta_i$ be the parameters for $i$-th sample, the regression loss is defined as:  
\begin{equation}
{\cal L}_{\rm regress} = \sum_{i} \; {\rm smooth}_{L1} (\theta_i - {\theta}_i^{\rm anno} )  \nonumber
\end{equation}
where  ${\theta}_i^{\rm anno}$ is the annotation provided from the registration step. Here we use the smooth L1 loss because of its robustness to outliers. This choice was more effective than the L2 loss in contributing to decreasing the error during the iterative training strategy in the presence of potential outliers and noisy annotations.

The body model is similar to the one we used for registration, except for  the pose representation, where we found that the use of quaternions improved stability and convergence of training than axis angle, which is probably due to the fact that the values of quaternions are in between -1 and 1 and is easier for ConvNets to learn with than axis angles.  Other parameters are same as the ones used in Section \ref{sec:regist}, which results in 132 parameters in total. Note that the global rotation is regressed using 9 parameters and the Gram Schmidt orthogonalization is used to make a transformation into a rotation. We use ResNet50 \cite{DBLP:journals/corr/HeZRS15} pretrained on the ImageNet dataset as the base network. 

\subsection{Inference and final refinement based on registration}

During the inference phase, there are two steps: 3D body parameter prediction and skeleton-based deformation. Since body shape/pose parameters are highly non-linear and are difficult to regress and predict accurately using deep ConvNets, we optionally provide a way to refine ConvNet predictions. This is based on the deformable registration technique proposed in Section \ref{sec:regist}. In order to define the dense correspondence term, we use DensePose \cite{Guler2018DensePose} to obtain dense uv maps and part indices (Fig. \ref{fig:refine_method}), which are then converted to image-surface correspondences. In addition, the simple baseline 2D human pose detector \cite{xiao2018simple} is used to obtain 2D human joint positions from an image and to define the key point loss. The pre-trained models from \cite{Guler2018DensePose} and \cite{xiao2018simple} are used.

	\section{Experimental results}
\subsection{Implementation and training detail}

Our method is implemented using Pytorch. Training takes 2-3 days using a NVIDIA Quadro P6000 graphics card with 24 GB memory. We use the Adam optimizer for all the steps in our approach. The multi-view cGANs is trained for 60 epochs with the batch size of 1024 and the learning rate of 0.0002. At each iteration, the body regressor is trained for 50 epochs with the batch size of 30 and the learning late of 0.0001. From the 1st to 4th iteration of training, we used both Human3.6M dataset and MS COCO dataset. At the last iteration (5th), we fine-tune the network on Human3.6M dataset only. We set the parameters in the loss function to $\alpha = \gamma =1$ and $\beta = 10$. For deformable surface registration, we use the learning rate of 0.1 and batch size of 10. We empirically set the parameters to $\omega_{\rm dense} = 1000$,  $\omega_{\rm KP} = 1$, $\omega_{\rm scale} = 10$, $\omega_{\rm joint}= 0.001$ and  $\omega_{\rm det} = 1$. For the first training iteration, in order to recover a global rotation, we set $\omega_{\rm scale} = 100$ and $\omega_{\rm joint}= 1$ to make the body model stiff, which is a common strategy in deformable registration \cite{ARV07}. We perform 300 forward-backward passes during the registration step at the 1st iteration. From the second iteration, 100 forward-backward passes were sufficient, since we start from the ConvNet prediction.        

\subsection{Dataset}
\begin{figure*}[tb]
	\centering
	\includegraphics[width=0.9\linewidth]{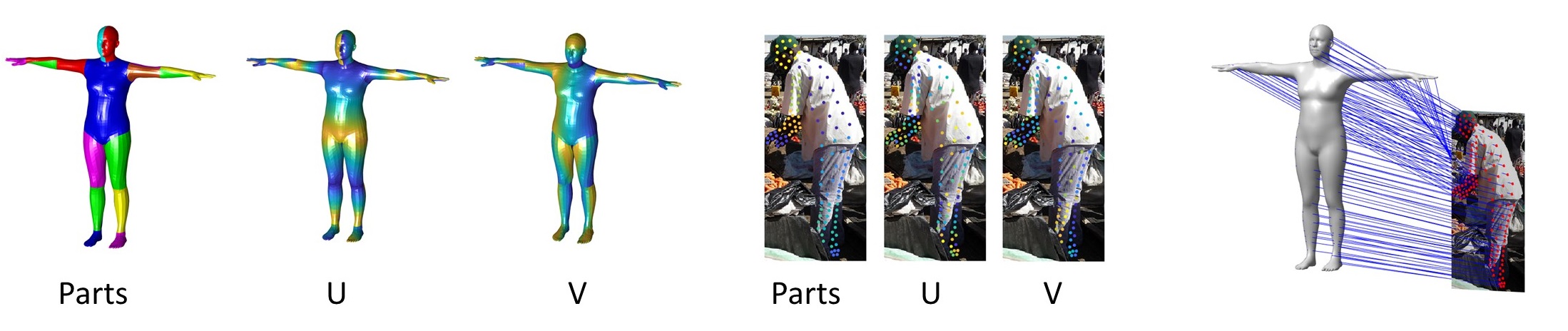}
	\caption{Dense image-surface correspondences between the template body surface and image points are found from DensePose annotations by searching nearest points in the UV  space of each body part. }
	\label{fig:dennse_annotation}
\end{figure*}

\noindent {\bf DensePose } DensePose dataset \cite{Guler2018DensePose} contains images with dense annotations of part-specific UV coordinates (Fig. \ref{fig:dennse_annotation}), which are provided on the MS COCO images. To obtain part-specific UV coordinates, body surfaces of a SMPL human body model are partitioned into 24 regions and each of them are unwrapped so that vertices have UV coordinates. Thus, every vertex on the model have unique parameterizations. Based on this, images are manually annotated with part indices and UV coordinates to establish dense image-to surface correspondences. To use this dense correspondences in 3D model fitting, we find the closest points from image pixels to surface vertices in UV coordinates of every part. The nearest neighbor search is done in this direction because image pixels are usually coarser than surface vertices. We were able to obtain approximately 15k annotated training images with the whole body contained in the image region.  

\noindent {\bf Human3.6M } Human 3.6M dataset  is a large scale dataset \cite{h36m_pami} for 3D human pose detection. This dataset contains 3.6 million images of 15 everyday activities, such as walking, sitting and making a phone call, which is performed by 7 professional actors and is taken from four different views. 3D positions of joint locations captured by MoCap systems are also available in the dataset. In addition, 2D projections of those 3D joint locations into images are available. To obtain dense annotations for this dataset, we use Mosh \cite{Loper:SIGASIA:2014} to obtain SMPL body and pose parameters from the raw 3D Mocap markers and then projected mesh vertices onto images to get dense correspondences between images and a template mesh. Note that some of the results are not well-aligned to markers and camera coordinates, resulting in a training dataset containing around $17k$ images and dense correspondence annotations. 

\noindent {\bf MPII 2D human pose } The images from MPII 2D human pose dataset \cite{andriluka14cvpr} is used for testing and was not used in training. Also, 2D keypoint labels in this dataset were used to trained the cGANs.   

\subsection{Protocol and metric}

We followed the same evaluation protocol used in previous approaches \cite{PavlakosZDD16,zhou2017weakly} for evaluation on Human3.6M dataset, where we use 5 subjects (S1, S5, S6, S7, S8) for training and the rest 2 subjects (S9, S11) for testing. The error metric for evaluating 3D joint positions is called mean per joint position error (MPJPE) in $mm$. Following \cite{zhou2017weakly} the output joint positions from ConvNets is scaled so that the sum of all 3D bone lengths is equal to that of a canonical average skeleton. 

We also evaluate the fit of the body model to images based on the mean per pixel error and mean per vertex error which measures distances from the ground truth to the predicted vertices in 2D image space and 3D space. Prior to calculate per-vertex error, we obtain a 
similarity transformation by Procrustes analysis  and align the predicted vertices to the ground truth; this is similar to the reconstruction error in the 3D joint estimation.

\subsection{Qualitative results}
\begin{figure*}[t]
	\centering
	\includegraphics[width=0.9\linewidth]{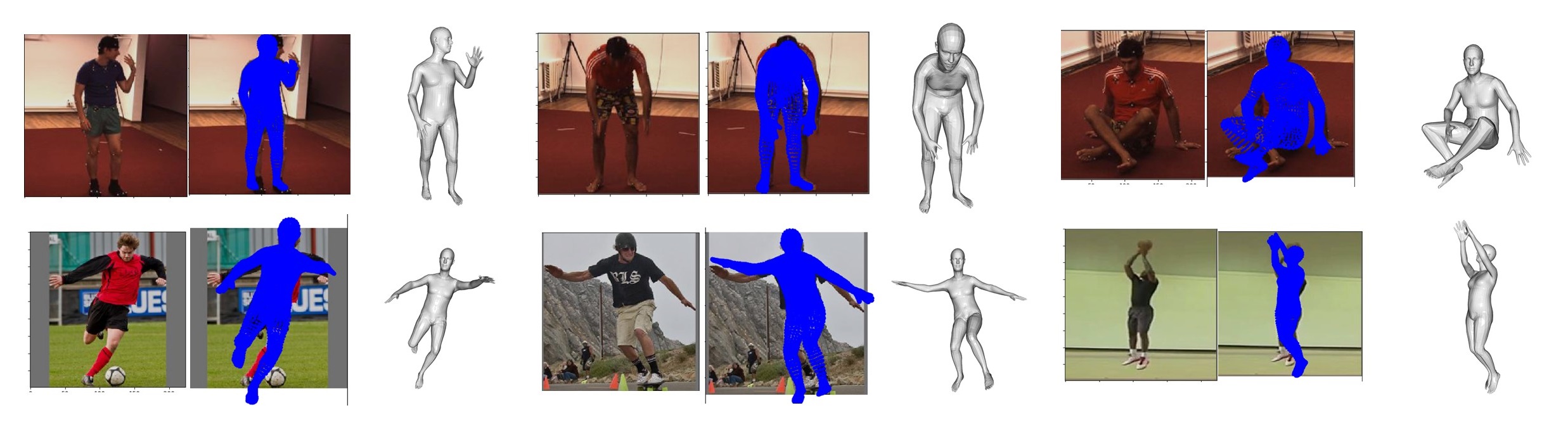}
	\caption{Qualitative results before refinement. Our technique is able to recover body shape and pose even from in-the wild images. }
	\label{fig:Qualitative}
\end{figure*}

\begin{figure*}[t]
	\centering
	\includegraphics[width=0.9\linewidth]{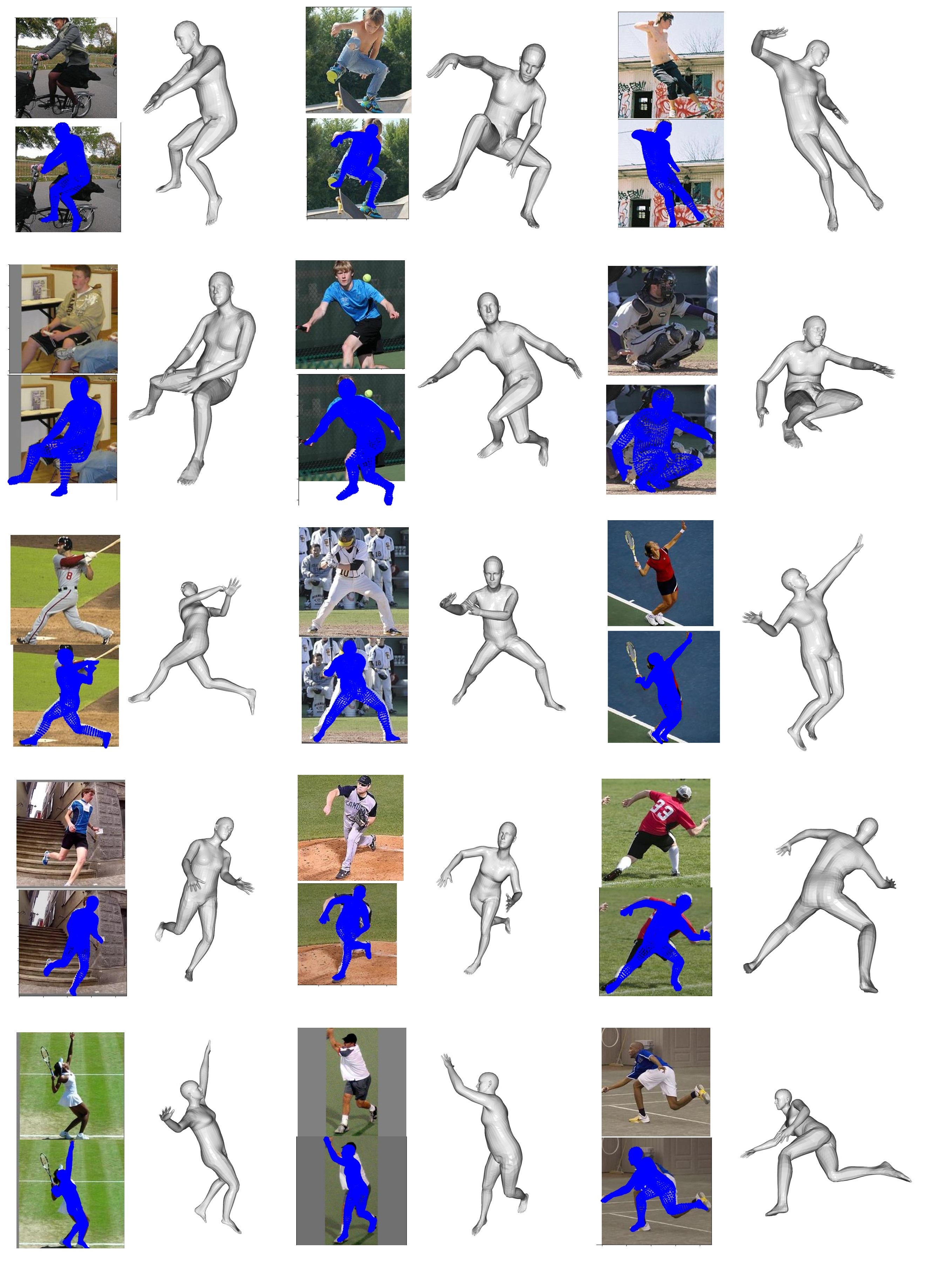}
	\caption{More results after refinement. }
	\label{fig:refinment2}
	
	\end{figure*}

In Figs. \ref{fig:Qualitative}, \ref{fig:refinment} and \ref{fig:refinment2}, we show our results on body shape and pose estimation before and after refinement. As we can see from the figures, our technique can predict 3D body shape and pose from in-the-wild images. Before refinement, the predicted poses are close to the images but still there are misalignments  especially at hands and feet. After refinement, the mesh is attracted toward image points based on dense correspondence predictions.

\begin{figure}[tb]
	\centering
	\includegraphics[width=1\linewidth]{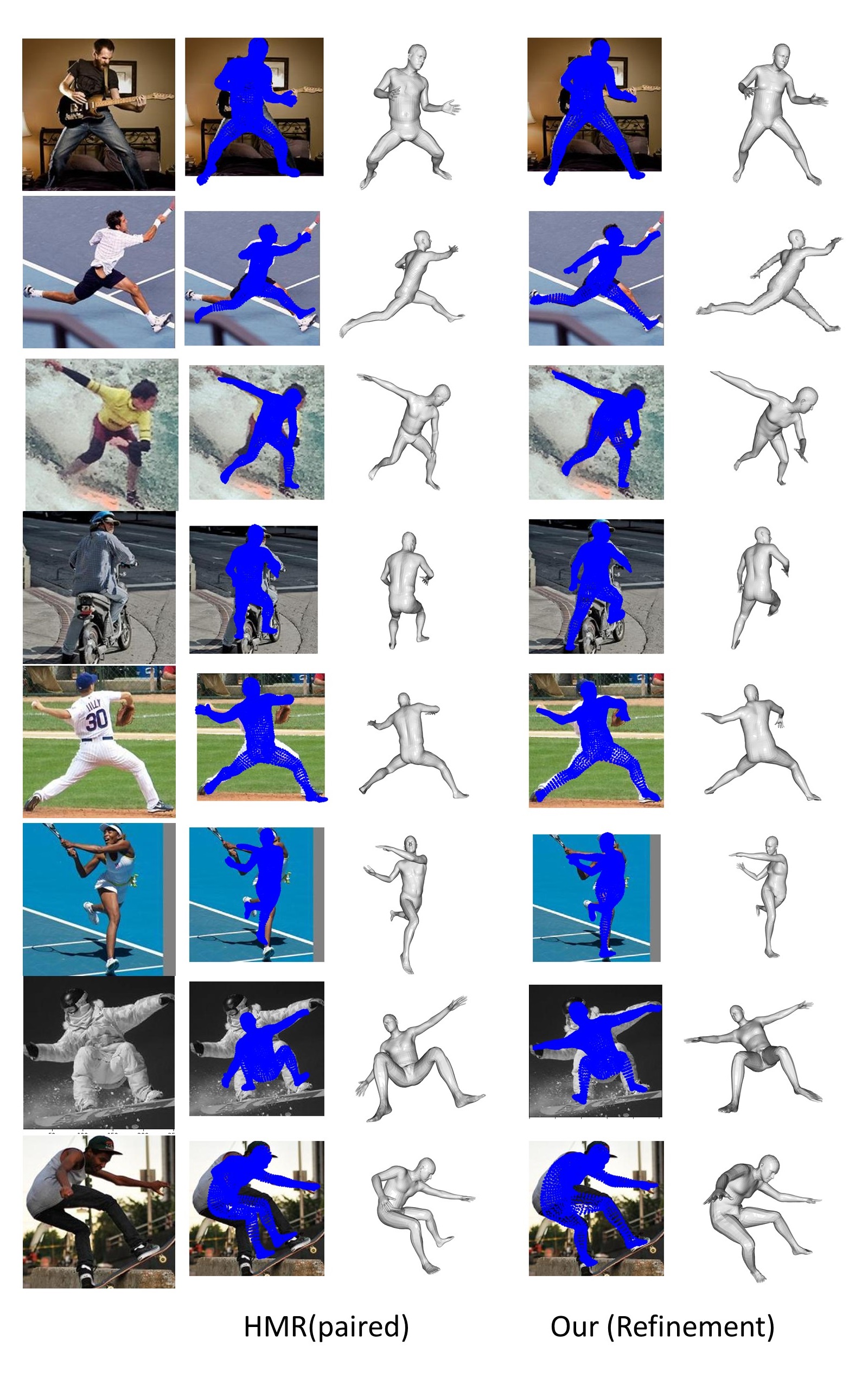}
	\caption{Comparisons to HMR \cite{hmrKanazawa17}. }
	\label{fig:comp_hmr}
\end{figure}

\subsection{Comparison with state-of-the-art}

\noindent {\bf 3D joint position and rotation } We compared our method with state-of-the-art techniques (Table 	\ref{tab:comparison2}). Here we only deal with unsupervised or weakly-supervised techniques which do not use full 3D supervisions. Kudo et al. \cite{kudo2018} uses conditional GANs to predict depths from 2D joint coordinates, which is the only technique that learns a model from 2D information only, except for ours. Rhodin et al. \cite{rhodin2018unsupervised} use an auto-encoder to compress visual features and reconstruct 3D pose from it, which does not require a large amount of 3D human poses. Our technique outperforms them in terms of MPJPE accuracy and is able to not only predict joints but also the orientations of limb segments as well as body shape represented in the form segment scales. Note that our 3D human pose cGANs even outperforms \cite{kudo2018} by incorporating geometric constraints. Our method is on par with HMR (unpaired) that uses 3D pose and body shape dataset for training GANs to provide 3D constraints in an unsupervised learning manner. On the other hand, we need  dense annotations on 2D images and do not use any 3D annotations such as body shape and pose. Note that HMR (paired) further provides images paired with 3D poses to do supervised learning, which makes their method  slightly better than ours but this requires an experimental setup with a Mocap system and synchronized video cameras to construct dataset.     

\begin{table*}[hbt]
	\begin{center}
		\caption{Comparisons with state of the art. MPJPE [mm] is used for error metric. }
		\label{tab:comparison2}
		\begin{tabular}{c c c c c c c}
			\hline 
			Kudo et al. \cite{kudo2018} & Rhodin et al. \cite{rhodin2018unsupervised} & HMR \cite{hmrKanazawa17} & HMR (paired) & Ours & Our cGANs & Ours (refine) \\ 
			\hline 
			173.2 & 131.7 &  106.84 & 87.97 & 106.25 & 139.9 & 108.46 \\ 
			\hline 
		\end{tabular}
	\end{center}
\end{table*}

\noindent {\bf Per-pixel and per-vertex error } In order to evaluate alignment of a body model to images, we measured the mean per-vertex error and mean per-pixel error and compare with HMR (paired),  which is shown in Table \ref{tab:pixelerr}. HMR (paired) obtained better results on Human 3.6M dataset than ours in both vertex alignment and pixel alignment, as they use a large amount of 3D pose dataset paired with images whereas ours only use 2D annotations. For this dataset, our refinement was not very effective probably because there is no large  variations of the subjects, actions and background in this dataset. For MS COCO dataset, our refinement was effective because this dataset is challenging for deep ConvNets to predict body parameters from due to a large variations in background, body shape/pose and clothing. Thanks to the refinement step, we achieve better fits in terms of the mean per-pixel error than HMR. Figure. \ref{fig:comp_hmr} shows the comparisons between ours and HMR (paired). Our method with refinement produces better alignment than HMR, especially around feet and hands. Our method captures more natural appearances of body shape and pose as the prior and constraints used do not come from the 3D dataset that is limited to some age range or captured in a controlled environment.

\begin{table*}[hbt]
\begin{center}
\caption{Comparisons to HMR (paired) in terms of per-pixel error and per-vertex error}
\label{tab:pixelerr}
\begin{tabular}{c c c c }
\hline 
&HMR (paired)  & Ours & Ours (refine) \\ 
\hline 
COCO per-pixel err. [pixel] & 13.9 & 18.6  & {\bf 12.02} \\ 
H3.6M per-pixel err. [pixel] & {\bf 7.3} &  9.9 &  9.2\\ 
H3.6M per-vertex recon. err. [mm] & {\bf 75.0}  & 102.7 & 97.2 \\
\hline 
\end{tabular}
\end{center}
\end{table*}

\begin{figure*}[tb]
	\centering
	\includegraphics[width=0.9\linewidth]{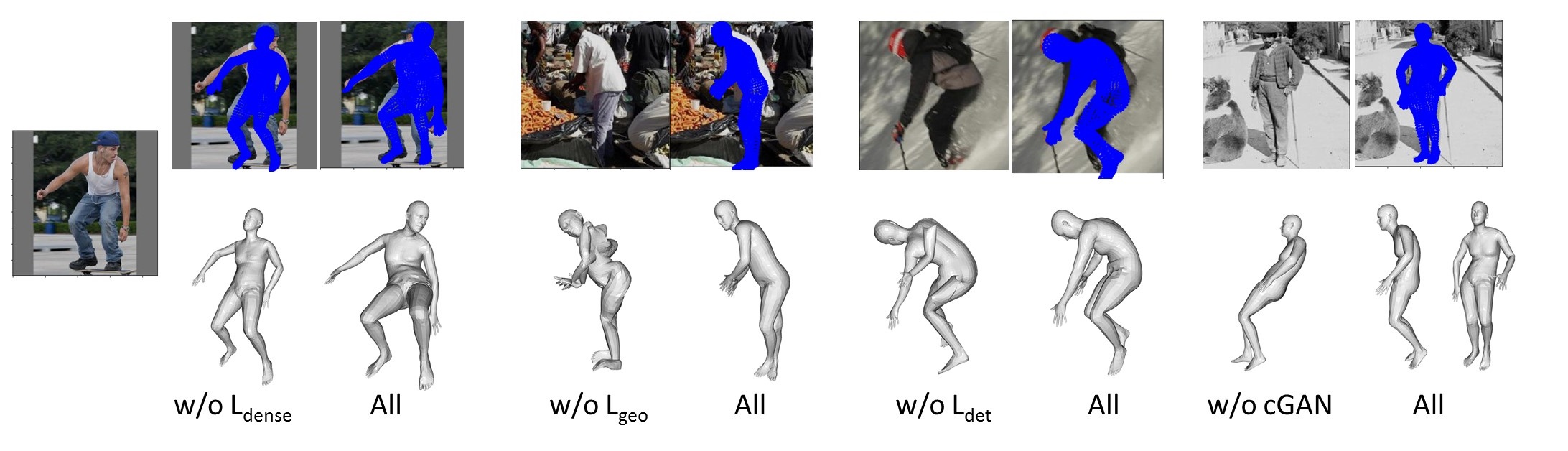}
	\caption{Ablation studies on image-surface registration. Without $\cal{L}_{\rm dense}$, the alignment to the image is poor, even the pose fit is not satisfactory. The losses $\cal{L}_{\rm geo}$ and $\cal{L}_{\rm det}$ play important role in mitigating distortions. Using cGANs to obtain depths at joints and provide soft constraints through $\cal{L}_{\rm KP}$, we can keep the depths at the joints statistically valid, which can for example prevent the incline or recline of the body model. }
	\label{fig:registration_results}
\end{figure*}

\begin{table}[hbt]
	\begin{center}
		\caption{Comparisons of MPJPE [mm] between training dataset. Note that the first iteration results are compared. }
		\label{tab:dataset}
		\begin{tabular}{c c c}
			\hline 
			MS COCO & Human3.6M & Both \\ 
			\hline 
			181.7 & 147.4 & 137.5 \\ 
			\hline 
		\end{tabular}
	\end{center}
\end{table}

\subsection{Ablation studies}
\noindent {\bf Loss } We have compared the registration results by varying the losses (Fig. \ref{fig:registration_results}). Without $\cal{L}_{\rm dense}$, the alignment between the body model and the image is poor, even the pose fit is not satisfactory. The losses $\cal{L}_{\rm geo}$ and $\cal{L}_{\rm det}$ play important role in mitigating distortions. With our 3D human pose cGANs, the depths of a  skeleton can be constrained by making the resulting distribution close to that of the data, which can for example prevent the incline and recline of the body.  

\noindent {\bf Dataset } We also compared the results of ConvNets  by changing the dataset i.e., using dense COCO only, Human3.6M only and both. The MPJPE results after the 1st iteration are shown in Table \ref{tab:dataset}. By combining MS COCO which contains in-the-wild images and Human3.6M dataset which includes domain knowledge, the better results are obtained than using a single dataset.

\begin{figure}[htbp]
	\begin{center}
		\includegraphics[width=80mm]{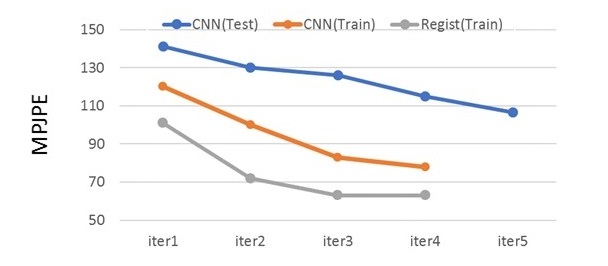}
	\end{center}
	\caption{History of MPJPE with respect to the number of iterations. Blue: MPJPE of ConvNet predictions on testing images; Orange: MPJPE of ConvNet predictions for training images; Gray: MPJPE evaluations of registration results on training dataset; From the first to fourth iteration, the network is trained using both MS COCO and Human 3.6M images. The fifth iteration fine-tunes the network with Human3.6M dataset only. }
	\label{fig:history}
\end{figure}

\begin{figure*}[htbp]
	\begin{center}
		\includegraphics[width = 0.85\linewidth]{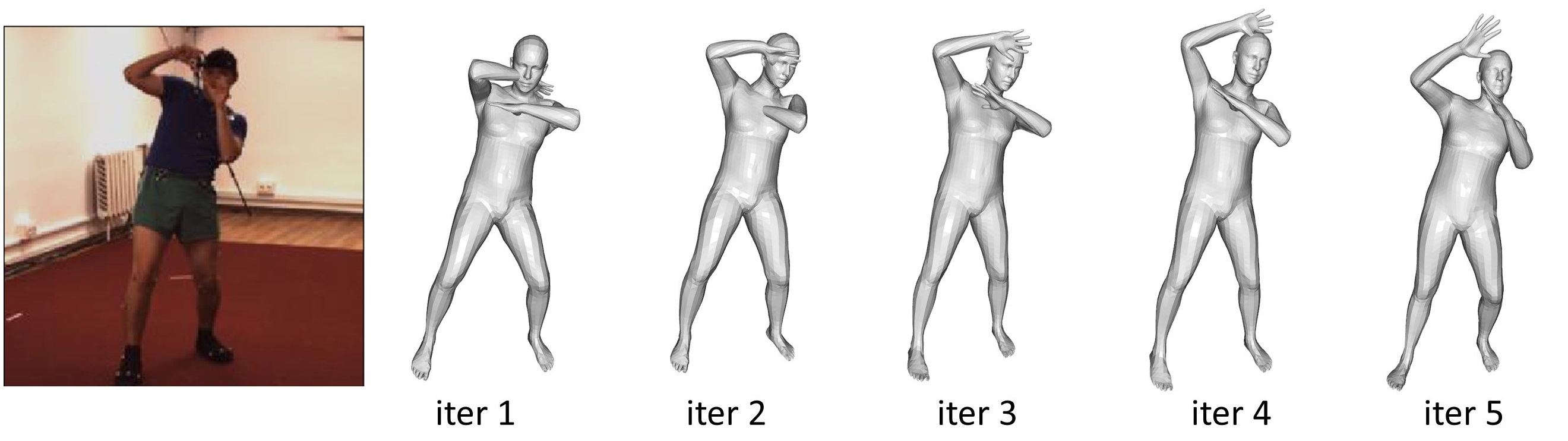}
	\end{center}
	\caption{Results after each iteration. As the number of training iterations increases, the body model fits to images better, which visualizes the effectiveness of iterative training. }
	\label{fig:vis_iterative}
\end{figure*}

\subsection{Is the iterative training strategy effective?}
To show the effectiveness of our iterative training strategy, we show a graph with the history of MPJPE errors in Fig. \ref{fig:history}. Here, MPJPE values after deformable registration is computed for training dataset. Our deform-and-learn strategy starts from image-surface registration using the T-pose as the initial pose. After the first registration phase, the train-set MPJPE for registration results is approx. 100 mm. Then, ConvNets is trained based on these registration results as supervisions. After 1 iteration, the test-set MPJPE of ConvNet predictions is 140 mm, which is slightly high. Next, deformable surface registration is performed again using the results of ConvNets as its initialization. These two steps are iterated for several times. This strategy was shown to be effective in gradually decreasing the error, which is visually noticeable in Fig. \ref{fig:vis_iterative}. In fact, MPJPE decreased approximately 30 mm from around 140 mm to 106 mm.  

We also compared different training strategies in Table \ref{tab:strategy}.  For the single-step learning strategy which incorporates all the losses from Eq. (\ref{eq:loss}) for training ConvNets (including the training of a discriminator), this is a difficult problem and the error stayed high. By omitting a discriminator and the loss for joint angle regularization ${\cal L}_{\rm joint}$, we were able to train the model  but the MPJPE error was not satisfactory. Also, instead of iterating registration and learning we have tried to perform one iteration of deform-and-learn and train longer (200 epoch). This improved MPJPE slightly but not as much as five iterations of deform-and-learn. Note also that a longer deformable registration (600 iterations) in a single step only improves accuracy of MPJPE 10mm (the train-set MPJPE from 100mm to 90mm), which is inferior to our deform-and-learn strategy that can achieve the train-set MPJPE 60 mm after registration.       
 
\begin{table*}[hbt]
	\begin{center}
		\caption{Comparisons between training strategies. MPJPE [mm] is used for error metric. }
		\label{tab:strategy}
		\begin{tabular}{c c c c}
			\hline 
			Single step (Eq. (\ref{eq:loss})) & Single step (w/o $\cal{L}_{\rm adv}$ \& $\cal{L}_{\rm joint}$)  & def-learn (1 iter 200 epoch) & def-learn (5 iter) \\ 
			\hline 
			 n/a & 148.1 & 134.8 & 106.25 \\ 
			\hline 
		\end{tabular}
	\end{center}
\end{table*}

\subsection{Inference time}

We measure the time required for the inference phase which can be divided in to five major steps: 3D body parameter prediction, skeleton-based deformation, 2D key point prediction (optional), DensePose prediction (optional) and refinement (optional). The 3D body parameter prediction step itself only takes approx. 0.035 sec.  The time for the skeleton deformation step is also approx. 0.035 sec, which means that the inference can be performed in approximately 0.07 sec given the cropped image of a human. Other steps that are required for refinement take 0.03 sec and 0.08 sec for 2D joint prediction and DensePose prediction, respectively. The refinement step takes 5-6 seconds for 50 iterations. Adding up all the step, our technique including refinement takes around 5 seconds to process one image. Compared to SMPLify \cite{bogo2016keep} and its variants \cite{Lassner:UP:2017}, which takes over 30 sec, our technique is faster as we start from the better initial pose and shape. 

\subsection{Failure cases and limitations}
\begin{figure*}[h]
		\centering
		\includegraphics[width=0.95\linewidth]{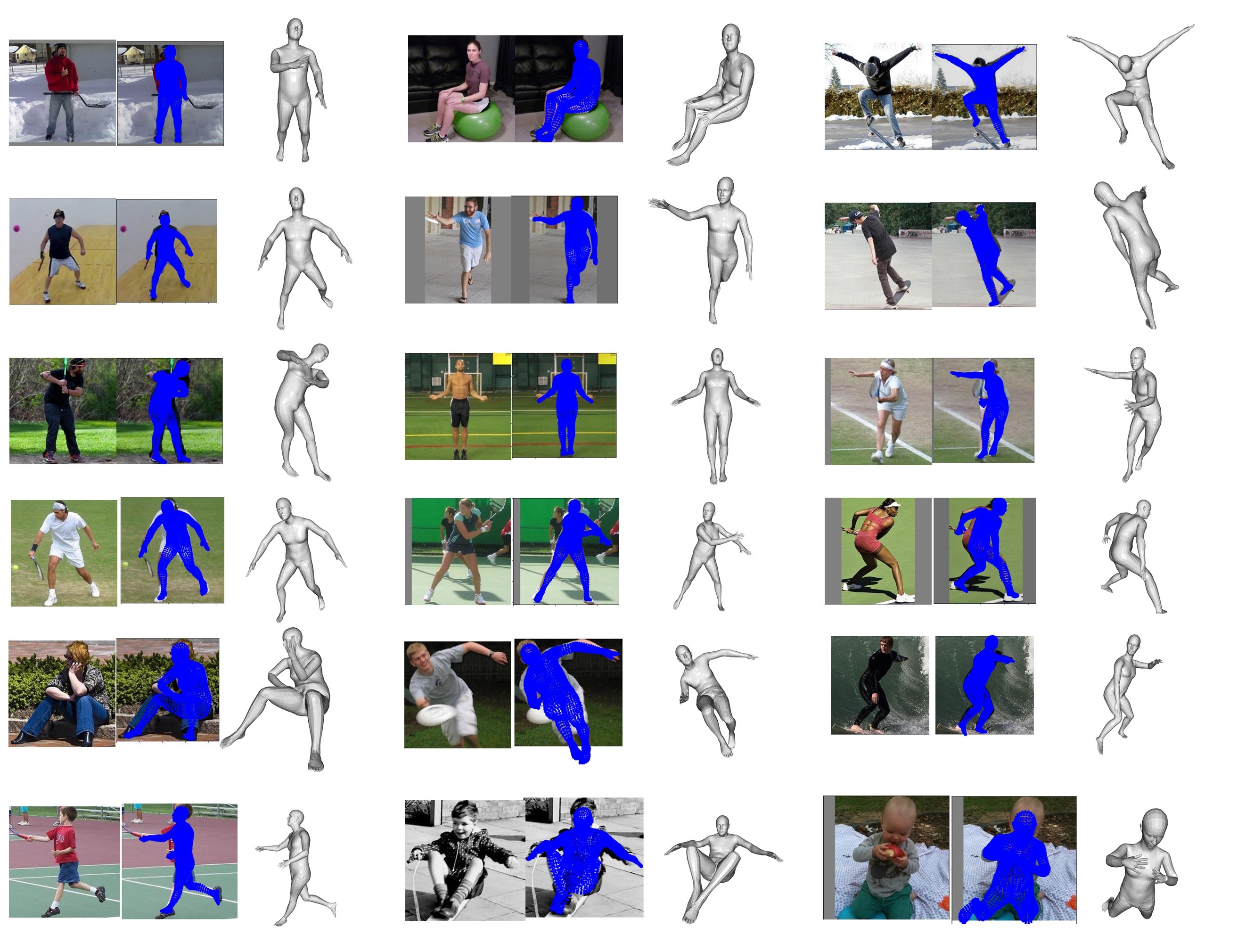}
		\caption{Results after refinement. Our technique is able to recover body shape and pose even from in-the wild images. }
		\label{fig:refinment}

\end{figure*}

As our refinement step rely on DensePose predictions, if this result is erroneousness, the final result will be worse; for example DensePose occasionally confuses the left and right for hands and feet, which results in distortions. While we represent body shape by segment scales, it is difficult to estimate a child's shape (Fig. \ref{fig:refinment}) as the body style is very different from the template mesh. A mechanism to select template meshes from different body styles would be useful for these cases. As with most of other approaches, our method cannot recover the absolute scale of body shape. Our network is currently designed for estimating body shape and pose for a single-person and we would like to extend to multiple human settings.

	\section{Conclusion}

We presented a deep learning technique for estimating 3D human body shape and pose from a single color image. To that end, we propose an iterative training approach that alternates between deformable surface registration and training of deep ConvNets, which gradually improves accuracy of predictions by extracting and aggregating 3D information from dense correspondences provided on 2D images. This approach allows us to learn 3D body shapes and pose from 2D dataset only without having to use 3D annotations that are in general very expensive to obtain. More importantly, as our approach does not rely on statistical body models or 3D annotations  captured in a controlled environment, our method is not restricted to the space of the pre-built statistical model and can capture body shape and pose details contained in in-the-wild images better than previous approaches. 

In future work, we would like to address the modeling of clothing and details. We are interested in designing a single unified network that can handle from 2D detection to 3D body shape/pose prediction all at once, which would be more efficient and faster. It would also be beneficial to find a better representation of 3D human body and pose than  body shape/pose parametric representation, which is more friendly to be used by ConvNets.


	\bibliographystyle{eg-alpha}
	\bibliography{egbib}

	

\end{document}